\documentclass{article}
\usepackage{amsfonts}
\usepackage{amsthm,amssymb,mathrsfs}
\usepackage{spconf,amsmath,graphicx}
\usepackage{graphicx}
\usepackage{multirow}
\usepackage{subfigure}

\title{Decoupled Non-parametric Knowledge Distillation for End-to-end Speech Translation}
\name{Hao Zhang$^1$, Nianwen Si$^{1,2}$, Yaqi Chen$^1$, Wenlin Zhang$^1$, Xukui Yang$^{1*}$, Dan Qu$^{1*}$, Zhen Li$^1$\thanks{*Corresponding Author. This work was supported by the National Natural Science Foundation of China under Grants 62171470.}}
\address{$^1$University of Information Engineering, Zhengzhou, China \\
$^2$Department of Electronic Engineering, Tsinghua University, Beijing, China}

\begin{document}
%\ninept
%
\maketitle
\begin{abstract}
Existing techniques often attempt to make knowledge transfer from a powerful machine translation (MT) to speech translation (ST) model with some elaborate techniques, which often requires transcription as extra input during training. However, transcriptions are not always available, and how to improve the ST model performance without transcription, i.e., data efficiency, has rarely been studied in the literature. In this paper, we propose Decoupled Non-parametric Knowledge Distillation (DNKD) from data perspective to improve the data efficiency. Our method follows the knowledge distillation paradigm. However, instead of obtaining the teacher distribution from a sophisticated MT model, we construct it from a non-parametric datastore via \emph{k}-Nearest-Neighbor (\emph{k}NN) retrieval, which removes the dependence on transcription and MT model. Then we decouple the classic knowledge distillation loss into target and non-target distillation to enhance the effect of the knowledge among non-target logits, which is the prominent “dark knowledge”. Experiments on MuST-C corpus show that, the proposed method can achieve consistent improvement over the strong baseline without requiring any transcription. 
\end{abstract}
\section{Introduction}
\label{sec:intro}

Different from the traditional cascading method \cite{dalmia2021searchable,bahar2021tight} which decomposes speech translation (ST) into two sub-tasks – automatic speech recognition (ASR) for transcription and machine translation (MT) for translation, end-to-end speech translation (E2E-ST) \cite{berard2018end,dong2021listen,fang2022stemm} aims at translating speech in source language into text in target language without generating transcription in source language. And the E2E-ST model has attracted more attention due to its advantages over the cascade paradigm, such as low latency, alleviation of error propagation and fewer parameters \cite{liu2020bridging}. However, since the translation alignment between speech and text is no longer subject to the monotonic assumption \cite{zhang2022revisiting}, it is non-trivial to train such a model well.

The high variation and fine granularity of speech make it more difficult to extract semantic information compared with text, resulting the ST model performance usually much inferior to the corresponding text-based MT model \cite{liu2020bridging}. Therefore, researchers often resort to a sophisticated MT model for cross-modal knowledge transfer with some techniques like knowledge distillation \cite{liu2019end}, contrastive learning \cite{han2021learning,ye2022cross}, triangular decomposition agreement \cite{du2021regularizing} and manifold mixup \cite{fang2022stemm}. Although those methods have achieved impressive performance improvements, they fail to take effect when transcription is not available. About 3000 languages in the world have no written form at all, for which it would be impractical to collect large amounts of phonetically transcribed data \cite{zhang2022revisiting}. Therefore, how to improve the performance of the E2E-ST model without approaching any transcription still remains a crucial issue. Recently, Zhang et al. \cite{zhang2022revisiting} tried to train the E2E-ST model from scratch without relying on transcriptions or any pretraining. However, they mainly focused on the network design especially Transformer, such as smaller vocabulary, deep post-LN encoder and wider feed-forward layer.

Inspired by the recent studies on the non-parametric methods for MT task \cite{khandelwal2020nearest,yang-etal-2022-nearest}, we propose Decoupled Non-parametric Knowledge Distillation (DNKD) from the data perspective to improve the data efficiency of E2E-ST task through distilling the knowledge of the non-parametric datastore, which is constructed from the baseline ST model and completely independent of transcription and MT model. DNKD yields an average improvement of 0.71 BLEU over the strong baseline on all 8 language pairs of MuST-C \cite{di2019must} dataset. Experiments show that it is possible to achieve significant improvement without approaching any transcription. And its availability is also greatly expanded since DNKD is model-agnostic.

\begin{figure*}[htb]
\centering
\includegraphics[width=0.90\linewidth]{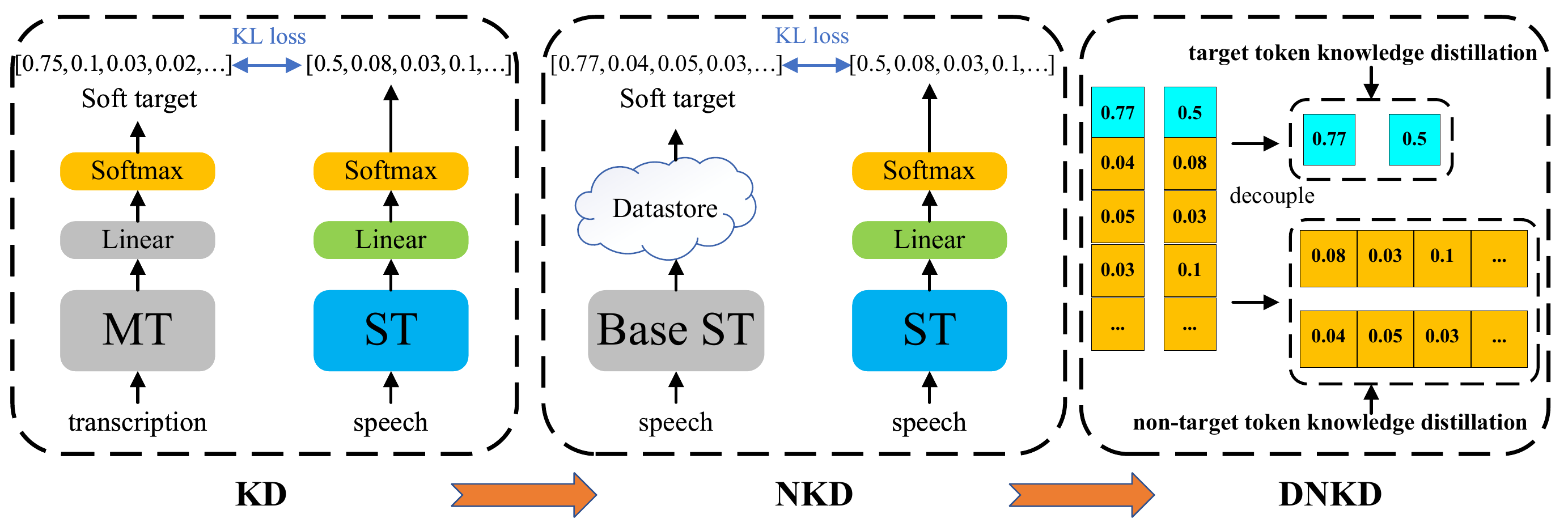}
\caption{Illustration of our idea. We first construct the teacher distribution from a non-parametric datastore which is built on the base ST model. In this way, we can reserve the knowledge transfer paradigm while remove the dependence on transcription and MT model in cross-modal KD (KD$ \to$NKD). Then we decouple the classical KL loss in knowledge distillation into target and non-target knowledge distillation to improve the transfer of knowledge contained in non-target logits, which is the prominent “dark knowledge” (NKD$ \to$DNKD).}
\label{fig:figure1}
\end{figure*}

\section{Methods}
\label{sec:pagestyle}

\subsection{Cross Modal Knowledge Distillation}
\label{section:1}
The ST corpus contains \emph{speech-transcription-translation} triples, denoted as ${\mathbf{ \cal D}} = \{ ({\emph{\textbf{s}}^{(n)}},{\emph{\textbf{x}}^{(n)}},{\emph{\textbf{y}}^{(n)}})\} _{n = 1}^N$, where $\emph{\textbf{s}}$ is the sequence of audio wave, $\emph{\textbf{x}}$ is the transcription in source language, and $\emph{\textbf{y}}$ is the translation in target language. Liu et al. \cite{liu2019end} first recommended cross-modal KD to improve the performance of ST model, in which MT and ST are termed as teacher and student model, respectively. The KD loss is defined as the KL-Divergence between teacher distribution ${\bf{p}}_{}^{\textbf{\rm{MT}}}$ and student distribution ${\bf p}_{}^{\textbf{\rm{ST}}}$.

\begin{equation}
{{{\cal L}}_{KD}} = \textbf{\rm{KL}}({\bf{p}}_{}^{\textbf{\rm{MT}}}||{\bf p}_{}^{\textbf{\rm{ST}}})
\label{equation:1}
\end{equation}

\subsection{Non-parametric Teacher Distribution Construction}
\label{section:2}
Although cross-modal KD has achieved impressive performance improvements on ST task, it relies on the transcription in source language as input to the well-trained MT model to obtain the teacher distribution. However, transcriptions are not always available as described above. Thus, instead of obtaining the teacher distribution from a parametric MT model, we construct it from a non-parametric datastore, which eliminates the need for transcription and MT model (KD$ \to$NKD). 

\noindent\textbf{Datastore creation} The datastore is constructed offline based on a trained baseline ST model and consists of a set of key-value pairs. Given the parallel \emph{speech-translation} pairs in training set ${\cal D}$, the datastore is created as follows:

\begin{equation}
({{\cal K}},{{\cal V}}) = \{ (f(\emph{\textbf{s}},\emph{\textbf{y}}_{ < i}^*),\emph{\textbf{y}}_i^*),\forall \emph{\textbf{y}}_i^* \in {\emph{\textbf{y}}^*}|(\emph{\textbf{s}},\emph{\textbf{y}}^*) \in {\cal D}\}
\end{equation}
where the key is the mapping high-dimensional representations of the translation context $(\emph{\textbf{s}},\emph{\textbf{y}}_{ < i}^*)$ in the training set using the mapping function $f$. Specially, we use the context vectors input to the final output layer as key \cite{khandelwal2020nearest,yang-etal-2022-nearest}. The value is the corresponding ground-truth $\emph{\textbf{y}}_i^*$. 
   
\begin{table*}[]
\centering
\begin{tabular}{l|lllllllll}
\hline
\multicolumn{1}{c|}{\multirow{2}{*}{\textbf{Models}}} & \multicolumn{9}{c}{\textbf{BLEU}}                                                                                              \\ \cline{2-10} 
\multicolumn{1}{c|}{}                        & En-De   & En-Fr   & En-Ru   & En-Es   & En-It   & En-Ro   & En-Pt   & \multicolumn{1}{l|}{En-Nl}   & Avg              \\ \hline
ESPnet \cite{inaguma_etal_2020_espnet}                                       & 22.91   & 32.76   & 15.75   & 27.96   & 23.75   & 21.90   & 28.01   & \multicolumn{1}{l|}{27.43}   & 25.06            \\
W2V2                                         & 23.86   & 34.74   & 15.98   & 28.84   & 24.48   & 22.56   & 29.10   & \multicolumn{1}{l|}{28.15}   & 25.96            \\
NKD                                          & 24.43** & 34.91*  & 16.14*  & 29.13*  & 24.81*  & 23.08** & 29.84** & \multicolumn{1}{l|}{28.55*}  & 26.36            \\
DNKD                                         & 24.79** & 35.22** & 16.42** & 29.45** & 25.12** & 23.31** & 30.23** & \multicolumn{1}{l|}{28.80**} & 26.67            \\ 
KD \cite{10042965}											 & 24.9    & 35.7	& 17.1		& 29.9	&	25.2	& 23.7 & 30.5 & \multicolumn{1}{l|}{29.6} & 27.07 \\ \hline
Datastore Size                               & 6.2M    & 7.8M    & 6.4M    & 5.9M    & 6.8M    & 6.1M    & 5.5M    & \multicolumn{1}{l|}{6.3M}    & \textbackslash{} \\ \hline
\end{tabular}
\caption{BLEU scores on MuST-C tst-COMMON set. W2V2 denotes our implemented baseline. NKD stands for training an ST model using knowledge distillation, where the teacher distribution is constructed from the datastore. DNKD further decouples target and non-target tokens on the basis of NKD. * and ** mean the improvements over W2V2 baseline is statistically significant ($p < 0.05$ and $p < 0.01$, respectively). KD represents the E2E-ST model trained by traditional cross-modal knowledge distillation.}
\label{fig:figure1}
\end{table*}

\noindent\textbf{Nearest neighbor search} Given a translation context $(\emph{\textbf{s}},\emph{\textbf{y}}_{ < i}^*)$ with ground-truth $\emph{\textbf{y}}_i^*$ in training set, we use $f(\emph{\textbf{s}},\emph{\textbf{y}}_{ < i}^*)$ to query the datastore to get the retrieved results ${{{\cal N}}^i}$ according to the squared-L2 distance, $d$.      

\begin{equation}
{{{\cal N}}^i} = \{ ({d_{ij}},{v_j}),j \in \{ 1,2, \cdots, k\} |(\emph{\textbf{key}}_j,{v_j}) \in ({{\cal K}},{{\cal V}}) \}
\end{equation}
where $\emph{\textbf{key}}_j$ is the \emph{j}-th nearest neighbor representation retrieved from the datastore, ${v_j}$ is the corresponding ground truth token, and ${d_{ij}} = d(\emph{\textbf{key}}_j,f(\emph{\textbf{s}},\emph{\textbf{y}}_{ < i}^*))$ is the squared-L2 distance between $\emph{\textbf{key}}_j$ and the query $f(\emph{\textbf{s}},\emph{\textbf{y}}_{ < i}^*)$. This is equivalent to extending the training data by adding $k$ reasonable target tokens for every translation context \cite{yang-etal-2022-nearest}.       
By conducting \emph{k}NN search in advance, the target sentence in the training set is extended from ${\emph{\textbf{y}}^*} = \{ y_1^*, \cdots ,y_n^*\} $ to ${\emph{\textbf{y}}^*} = \{ (y_1^*,{{{\cal V}}^1}), \cdots ,(y_n^*,{{{\cal V}}^n})\} $, where ${{{\cal V}}^i} = \{ {v_j},j \in \{ 1,2, \cdots, k\} \} $ is the value in ${{{\cal N}}^i}$. We use $\emph{\textbf{y}}_i^*$ and ${{{\cal V}}^i}$ to construct one-hot and teacher distribution, respectively.

\noindent\textbf{Teacher distribution construction} The squared-L2 distances, between the retrieved $k$ nearest neighbor representation and the query, are converted into the teacher distribution over the vocabulary by applying a softmax with temperature $\tau$ to the negative distances and aggregating over multiple occurrences of the same vocabulary item \cite{khandelwal2020nearest}. The teacher distribution is constructed as follows:

\begin{equation}
{\bf p}_{}^{k\textbf{\rm{NN}}}({\emph{\textbf{y}}_i}|\emph{\textbf{s}},\emph{\textbf{y}}_{ < i}^*) \propto \sum\limits_{({d_{ij}},{v_j}) \in {{{\cal N}}^i}} {{\mathbb{I}_{{\emph{\textbf{y}}_i} = {\emph{\textbf{v}}_j}}}\exp (\frac{{ - {d_{ij}}}}{\tau })}
\end{equation}
where $\mathbb{I}$ is indicator function. We use ${\bf{p}}_{}^{k\textbf{\rm{NN}}}$ as teacher to train the ST model. Equation~\ref{equation:1} can be rewritten as:

\begin{equation}
{{{\cal L}}_{NKD}} = \textbf{\rm{KL}}({\bf{p}}_{}^{k\textbf{\rm{NN}}}||{\bf p}_{}^{\textbf{\rm{ST}}})
\label{equation:5}
\end{equation}

\subsection{Decouple the target and non-target token}
\label{section:3}
Recent study has shown that the knowledge transfer of target and non-target class in classical KD loss is tightly coupled \cite{zhao2022decoupled}, which limits the effectiveness of knowledge distillation \cite{hinton2015distilling}. For each time step, the out probabilities distribution of the model is defined as $[{p_1},{p_2}, \cdots ,{p_v}] \in {\mathbb{R}^{1 \times V}}$, where ${p_i}$ is the probability of belonging to the \emph{i}-th token and $V$ is the size of vocabulary. We use target token to represent the ground-truth token in the output probability distribution at each time step, and other tokens are defined as non-target tokens. The binary probability of belonging to the target token ($p_t$) and all other non-target tokens ($p_{\backslash t}$) is defined as ${\bf b} = [{p_t},{p_{\backslash t}}] \in {\mathbb{R}^{1 \times 2}}$, where ${p_t} = \frac{{\exp ({z_t})}}{{\sum _{j = 1}^V\exp ({z_j})}}$ and ${p_{\backslash t}} = 1 - {p_t}$. ${z_j}$ is the logit of the \emph{j}-th token. The probability among non-target tokens is defined as ${\bf{\hat p}}=[{\hat p_1}, \cdots {\hat p_{t - 1}},{\hat p_{t + 1}} \cdots ,{\hat p_v}] \in {\mathbb{R}^{1 \times (V-1)}}$, where ${\hat p_i} = \frac{{\exp ({z_i})}}{{\sum _{j = 1,j \ne t}^V\exp ({z_j})}}$.  
The loss in Equation~\ref{equation:5} can be re-written as:

\begin{equation}
\resizebox{.9\hsize}{!}{$
{{{\cal L}}_{NKD}} = \textbf{\rm{KL}}({{\bf {b}}^{k\textbf{\rm{NN}}}}||{{\bf b}^{\textbf{\rm{ST}}}}) + (1 - p_t^{k\textbf{\rm{NN}}})\textbf{\rm{KL}}({{ \bf {\hat p}}^{k\textbf{\rm{NN}}}}||{{\bf {\hat p}}^{\textbf{\rm{ST}}}})
\label{equation:6}
$}
\end{equation}

where \emph{k}NN and ST are teahcer and student model, respectively. $p_t^{kNN}$ represents the teacher prediction on the target token. The first term is the similarity between the teacher and student distribution of the target token. The second term is the similarity among non-target tokens. The more confidence the teacher is in the training sample, the more reliable and valuable knowledge about non-target tokens it could provide. However, the loss weight of the second term is highly suppressed by the confident prediction $p_t^{kNN}$. Thus, we decouple the training objective in Equation~\ref{equation:6} as target and non-target token knowledge distillation to boost the transfer of knowledge among non-target tokens. During training, the ST model not only learn from the ground-truth, but also the soft target of the teacher model. The final training objective is as follows: 

\begin{equation}
\resizebox{.9\hsize}{!}{$
{{{\cal L}}_{all}} = \lambda {{{\cal L}}_{CE}} + (1 - \lambda )(\alpha \textbf{\rm{KL}}({{\bf {b}}^{k\textbf{\rm{NN}}}}||{{\bf b}^{\textbf{\rm{ST}}}}) + \beta \textbf{\rm{KL}}({{ \bf {\hat p}}^{k\textbf{\rm{NN}}}}||{{\bf {\hat p}}^{\textbf{\rm{ST}}}}))
$}
\label{equation:7}
\end{equation}

\section{Experiments}
\subsection{Datasets and pre-processing}

We conduct experiments on MuST-C dataset \cite{di2019must} to verify the effectiveness of DNKD. It contains translation from English (En) to 8 languages: German (De), French (Fr), Russian (Ru), Spanish (Es), Italian (It), Romanian (Ro), Portuguese (Pt), and Dutch (Nl). We use dev set for validation and tst-COMMON set for test.

For speech input, we use the raw 16-bit 16kHz mono-channel audio wave. We tokenize and truecase all text via Mose and use BPE \cite{sennrich2015neural} to learn a vocabulary on each translation direction. The vocabulary size is set to 8000.

\subsection{Model settings}
We use the Wav2vec 2.0 large model \cite{baevski2020wav2vec} pretrained on LibriSpeech \cite{7178964} audio data without finetuning to extract feature from raw speech as model input. We use Transformer \cite{vaswani2017attention} as our base model, which has 6 layers in both encoder and decoder. Each of these transformer layers comprises 256 hidden units, 4 attention heads, and 2048 feed-forward hidden units. We regular the model with label smoothing \cite{szegedy2016rethinking} and dropdim \cite{zhang2022dropdim} to prevent overfitting. The value of label smoothing is set to 0.1. We use the random mask strategy in dropdim with a mask rate 0.05. We train our models with Adam optimizer \cite{kingma2014adam} on 2 NVIDIA V100 GPUs. We use FAISS \cite{johnson2019billion} for the nearest neighbor search. We compute the case-sensitive BLEU scores and the statistical significance of translation results with paired bootstrap resampling \cite{koehn-2004-statistical} using scareBLEU \cite{post-2018-call}.

\subsection{Main Results}
Table 1 shows the main results. ESPnet \cite{inaguma_etal_2020_espnet} is a widely used ST system which uses fbank features as input and is equipped with pre-training and data augmentation. Our implemented W2V2 baseline takes features extracted by Wav2vec 2.0 as input. The comparation between W2V2 and ESPnet shows that W2V2 is a strong baseline. Without any transcription data, NKD achieves an average improvement of 0.4 BLEU over the strong baseline W2V2 by only learning from the non-parametric datastore. DNKD can further improve 0.3 BLEU over NKD through decoupling the knowledge distillation as target and non-target knowledge distillation. The performance of DNKD is also comparable to the conventional cross-modal KD (26.67 vs 27.07), illustrating the effectiveness of DNKD.

\subsection{Ablation Studies}

\begin{figure}[htb]
\subfigure{
\begin{minipage}[b]{0.3\linewidth}
  \centering
  \centerline{\includegraphics[width=1.0in]{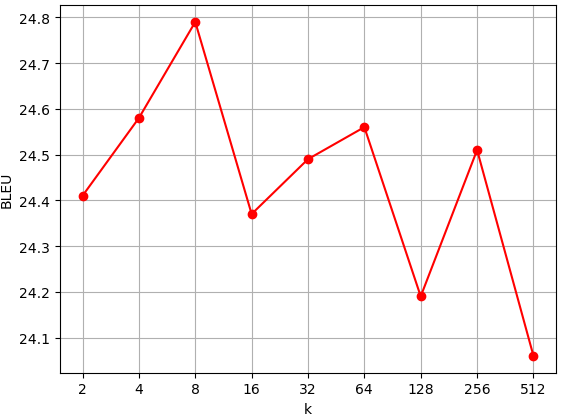}}
%  \vspace{1.5cm}
  \centerline{(a)}\medskip
\end{minipage}
}%
\subfigure{
\begin{minipage}[b]{0.3\linewidth}
  \centering
  \centerline{\includegraphics[width=1.0in]{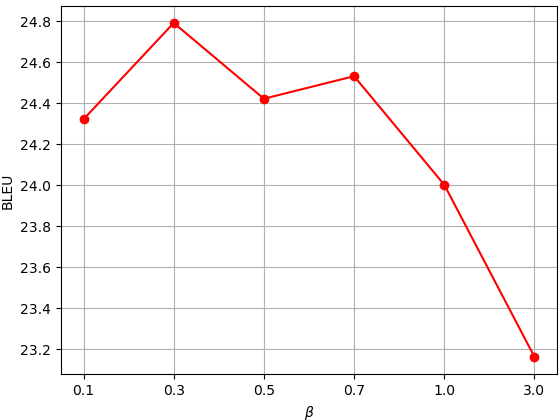}}
%  \vspace{1.5cm}
  \centerline{(b)}\medskip
\end{minipage}
}%
\subfigure{
\begin{minipage}[b]{0.3\linewidth}
  \centering
  \centerline{\includegraphics[width=1.0in]{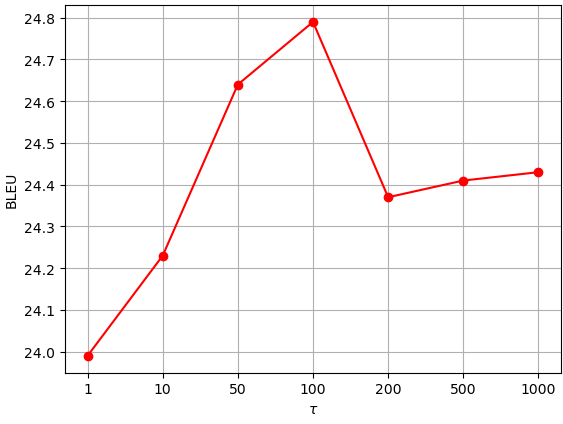}}
%  \vspace{1.5cm}
  \centerline{(c)}\medskip
\end{minipage}
}%
\caption{Ablation studies on the tst-COMMON set of MuST-C en-de dataset. (a) BLEU scores with different $k$ ($\tau  = 100,\beta  = 0.3$). (b) BLEU scores with different $\beta$ ($k = 8,\tau  = 100$). (c) BLEU scores with different $\tau$ ($k = 8,\beta  = 0.3$).}
\label{fig:2}
\end{figure}

\begin{figure}[h]
\begin{minipage}[b]{\linewidth}
  \centering
  \centerline{\includegraphics[scale=0.13]{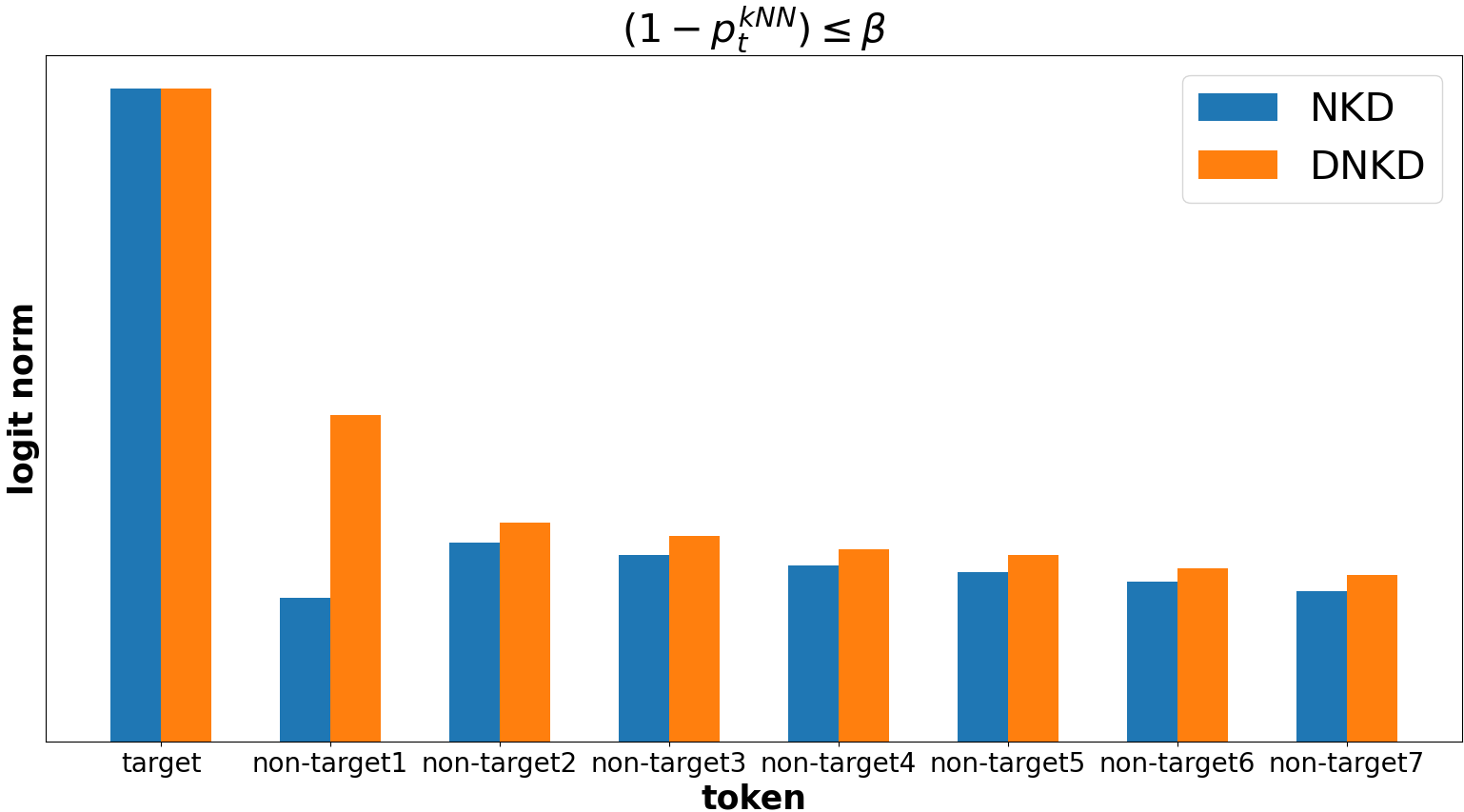}}
%  \vspace{1.5cm}
  \centerline{(a)}\medskip
\end{minipage}
%\hfill
\begin{minipage}[b]{\linewidth}
  \centering
  \centerline{\includegraphics[scale=0.13]{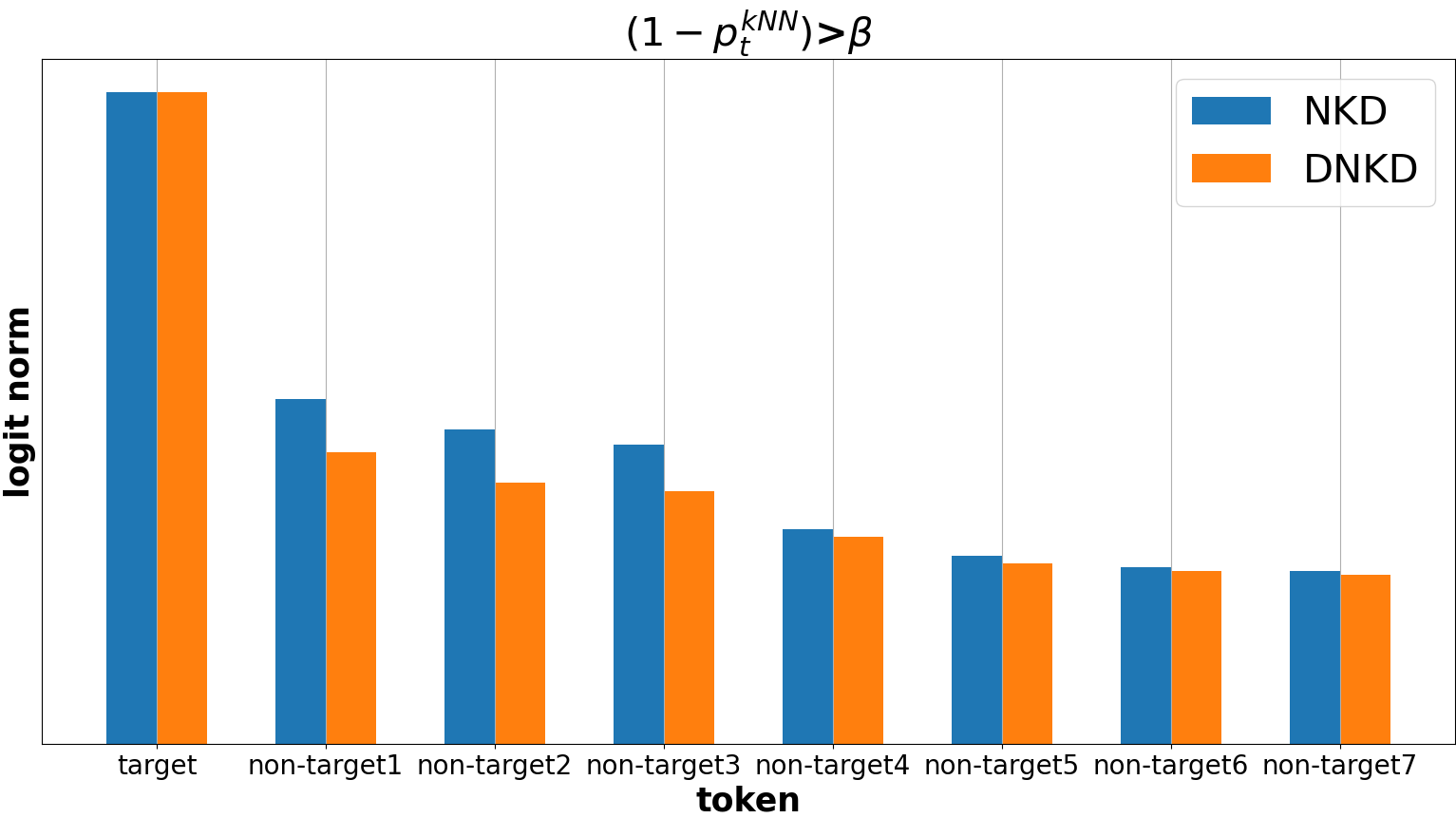}}
%  \vspace{0.0cm}
  \centerline{(b)}\medskip
\end{minipage}
\caption{(a) The top 8 gradient norm in the case of $(1 - p_t^{kNN}) \le \beta$. (b) The top 8 gradient norm in the case of  $(1 - p_t^{kNN}) > \beta$.}
\label{fig:3}
\end{figure}

$\lambda$ in Equation~\ref{equation:7} is set to 0.5 to make the learning process more robust. We mainly focus on the transfer of knowledge among non-target tokens, so $\alpha$ in Equation~\ref{equation:7} is set to default value 1.0. In this section, we mainly investigate the effect of another three key hyper-parameters.

\noindent\textbf{Effect of $k$} As shown in Figure~\ref{fig:2} (a), the BLEU value first rises with the increase of $k$, and achieves the peak of 24.79 when $k = 8$. The performance does not improve when retrieving a larger number of neighbors. This can attribute to that some retrieval tokens may not the reasonable target tokens when $k$ get larger, which will add more noise to model training.

\noindent\textbf{Effect of $\beta$} As shown in Figure~\ref{fig:2} (b), the model gets the best performance when $\beta  = 0.3$, and its performance does not improve when $\beta$ get larger. $\beta$ control the contribution of the non-target tokens in knowledge transfer. However, larger $\beta$ will also increase the gradient contributed by logits of non-target tokens. Thus, an improper large $\beta$ could harm the correctness of the ST model prediction.

\noindent\textbf{Effect of $\tau$} As shown in Figure~\ref{fig:2} (c), a temperature of 1 result in significantly lower BLEU scores than those greater 1, because large temperature value can flatten the \emph{k}NN teacher distribution to prevent assigning most of the probability mass to a single neighbor \cite{yang-etal-2022-nearest}. However, a temperature value that is too large can make the model less confident about the ground-truth. In our experiments, values of 100 prove to be optimal.

\subsection{Gradient Analysis}
In this section, we study the gradient norm over the logits to explain why the decoupling of target and non-target tokens is effective. The larger the value of the gradient norm, the more the model pays attention to the corresponding token \cite{yang-etal-2022-nearest,lin2021straight}. We analyze the following two cases: $(1 - p_t^{kNN}) \le \beta$ and $(1 - p_t^{kNN}) > \beta$.

In the case of $(1 - p_t^{kNN}) \le \beta$, DNKD will increase the non-target knowledge distillation scale factor from $(1 - p_t^{kNN})$ to $\beta$ through the decoupling method. The non-target knowledge distillation part in Equations~\ref{equation:7} does not produce gradient over the target token and we set $\alpha  = 1.0$ for the target knowledge distillation. Thus, as shown in Figure~\ref{fig:3} (a), DNKD can increase the gradient norm over non-target tokens logits while keeping the gradient norm over the target token logit. Figure~\ref{fig:3} (b) shows the results under the case of $(1 - p_t^{kNN}) > \beta$. In this situation, the small value of $p_t^{kNN}$ means that the teacher is less confident about its prediction of the target token. The knowledge about the non-target tokens is also unreliable. DNKD can reduce the gradient contributed by the logits of non-target tokens by reducing the non-target knowledge distillation scale factor from $(1 - p_t^{kNN})$ to $\beta$.

\section{Conclusion}
In this paper, we propose decoupled non-parametric knowledge distillation (DNKD) to improve the data efficiency of E2E-ST task. Experiments demonstrate the effectiveness of the proposed method. In the future, we are interested in exploring how the proposed techniques can advance the state-of-the-art methods which is coupled with transcription.
\bibliographystyle{IEEEbib}
\bibliography{refs}

\end{document}